\documentclass{article}

   \usepackage[final,nonatbib]{neurips_2024}

\usepackage[utf8]{inputenc} 
\usepackage[T1]{fontenc}    
\usepackage{url}            
\usepackage{booktabs}       
\usepackage{amsfonts}       
\usepackage{nicefrac}       
\usepackage{microtype}      
\usepackage{xcolor}         
\usepackage[utf8]{inputenc} 
\usepackage[T1]{fontenc}    
\usepackage{hyperref}       
\usepackage{url}            
\usepackage{booktabs}       
\usepackage{amsfonts}       
\usepackage{nicefrac}       
\usepackage{microtype}      
\usepackage{xcolor}         
\usepackage{makecell}
\usepackage[pdftex]{graphicx} 
\usepackage{multirow}
\usepackage{tabularx}
\usepackage{amssymb}
\usepackage{amsmath}
\usepackage{algorithm,algcompatible}
\usepackage{caption}
\usepackage{subcaption}
\usepackage{placeins}
\usepackage{cleveref}
\usepackage{algorithm}
\usepackage{amsmath}
\usepackage{booktabs}
\usepackage{makecell}
\usepackage{tabu, booktabs}
\usepackage{bm}
\usepackage{wrapfig}
\usepackage{tikz}
\usepackage{scalerel}
\usepackage{xspace}
\usepackage{pifont}
\usepackage{multirow}
\usepackage{tikz}
\usepackage{comment}
\usepackage{amsmath,amssymb} 
\usepackage{color}
\usepackage[inline]{enumitem}
\usepackage{float}
\usepackage{booktabs}       
\usepackage{indentfirst}
\usepackage{multirow}
\usepackage{makecell}
\usepackage{pifont}
\usepackage{everyshi}
\usepackage{colortbl}
\usepackage{marvosym}
\usepackage{newfloat}
\usepackage{listings}

\title{TabDeco: A Comprehensive Contrastive Framework for Decoupled Representations in Tabular Data }

\author{%
Suiyao Chen\\
Buyer Abuse Prevention\\
\texttt{suiyaoc@amazon.com} \\
\And
Jing Wu\\
WWPS Solution Architecture\\
\texttt{jingwua@amazon.com} \\
\And
Yunxiao Wang, \quad Cheng Ji, \quad Tianpei Xie \\
Buyer Abuse Prevention\\
\texttt{\{yunxiaow, cjiamzn, lukexie\}@amazon.com} \\
\And
Daniel Cociorva$^{1}$, Michael Sharps$^{2}$, Cecile Levasseur$^{1}$, Hakan Brunzell$^{1}$ \\
Buyer Abuse Prevention$^{1}$, WWPS Solution Architecture$^{2}$\\
\texttt{\{cociorva, sharpsm, cecilele, brunzell\}@amazon.com} \\
}

\begin{document}

\maketitle
\graphicspath{{./figures/}}

\begin{abstract}
Representation learning is a fundamental aspect of modern artificial intelligence, driving substantial improvements across diverse applications. While self-supervised contrastive learning has led to significant advancements in fields like computer vision and natural language processing, its adaptation to tabular data presents unique challenges. Traditional approaches often prioritize optimizing model architecture and loss functions but may overlook the crucial task of constructing meaningful positive and negative sample pairs from various perspectives like feature interactions, instance-level patterns and batch-specific contexts. To address these challenges, we introduce TabDeco, a novel method that leverages attention-based encoding strategies across both rows and columns and employs contrastive learning framework to effectively disentangle feature representations at multiple levels, including features, instances and data batches. With the innovative feature decoupling hierarchies, TabDeco consistently surpasses existing deep learning methods and leading gradient boosting algorithms, including XGBoost, CatBoost, and LightGBM, across various benchmark tasks, underscoring its effectiveness in advancing tabular data representation learning.
\end{abstract}


\section{Introduction}

Self-supervised learning methods, particularly contrastive learning, have gained popularity in response to the challenges of acquiring labeled data. This approach often rivals or even surpasses supervised learning, especially in computer vision and natural language processing \cite{wu2023hallucination,wu2023genco,wang2023balanced,lai2024residual,lai2024adaptive}, by creating embeddings that distinctly separate similar and dissimilar data points. In visual tasks, techniques like image rotation or puzzle assembly enhance positive pair similarities, while token masking in text processing helps capture robust, invariant features. However, several critical industries, including healthcare\cite{qayyum2020secure,chen2017personalized,chen2019claims,chen2024deep}, manufacturing\cite{borisov2022deep, chen2017multi,bingjie2023optimal,chen2020some}, agriculture\cite{liakos2018machine, wu2022optimizing, wu2024new,tao2022optimizing,wu2024crops,wu2023extended} and various engineering fields \cite{phatak2023computing,chen2018data,lahn2023combinatorial,raghvendra2024new}, still heavily rely on structured tabular data. 
Researchers traditionally leverage domain expertise for feature selection and uncertainty quantification \cite{ye2023demultiplexing,ye2023multiplexed,ye2024multiplexed,ye2023oam,ye2023free}.

However, when applied to tabular data, contrastive learning encounters unique challenges in constructing meaningful positive and negative samples. Traditional methods frequently alter individual features or use entirely different samples, lacking a deeper exploration of feature interactions, instance-level patterns and batch-specific contexts, which could be essential for enhancing the learning process.

To produce more structured representations for tabular data, one notable attempt is SwitchTab \cite{wu2024switchtab}, which employs an asymmetric encoder-decoder framework to decouple shared and unique features within data pairs. This feature decoupling process could naturally construct meaningful positive and negative samples, thus facilitating the contrastive learning process. However, this decoupling process faces notable difficulties. The linear projector used in SwitchTab often struggles to effectively delineate feature boundaries, leading to embeddings that are poorly organized and difficult to interpret, ultimately limiting the robustness of the approach. Another recent innovation, Self-Attention and Intersample Attention Transformer (SAINT) \cite{somepalli2021saint}, which has enhanced feature learning by integrating attention mechanisms at both column and row levels, however, may struggle with dataset complexity, especially when faced with high-dimensional, heterogeneous, and noisy data. 

The complementary strengths and limitations of SwitchTab and SAINT present an opportunity to develop a more robust framework for tabular data representation learning, focusing on effective feature decoupling to generate meaningful positive and negative sample pairs for contrastive learning, as well as capturing the data complexities to ehance model performance. Therefore, we propose TabDeco, an innovative contrastive learning framework that utilizes attention mechanisms to achieve finer-grained decoupling of local and global features across multiple levels in tabular data. TabDeco constructs positive and negative pairs from multiple perspectives, including local and global contrasts, feature-level and instance-level contrasts, enabling a more refined separation of feature hierarchies. By integrating attention-based encoding and feature decoupling strategies, TabDeco enhances the model’s capacity to isolate and emphasize relevant features and instances, surpassing the limitations of existing approaches and enhancing the representation learning performance.

Our contributions can be summarized as follows:
\begin{itemize}[noitemsep,topsep=0pt]

    \item[$\bullet$]We propose TabDeco, a novel contrastive learning framework for decoupled representation learning for tabular data. To the best of our knowledge, this is the first attempt to explore and explicitly facilitate structured embeddings learning through contrasting for tabular data. 

    \item[$\bullet$]By integrating the strengths from feature decoupling and attention-based learning, we demonstrate out method to achieve competitive results across extensive datasets and benchmarks.

    \item[$\bullet$]We develop the comprehensive framework for contrastive learning on decoupled features and construct positive and negative pairs from diverse perspectives, enhancing the exploration of feature interactions at local, global, and instance levels.

\end{itemize}

\section{Related Work}

\subsection{Classical vs Deep Learning Models}
For tasks like classification and regression in tabular data, traditional machine learning methods remain effective. Logistic Regression (LR) \cite{wright1995logistic} and Generalized Linear Models (GLM) \cite{hastie2017generalized} are commonly used for modeling linear relationships. In contrast, tree-based models such as Decision Trees (DT) \cite{breiman2017classification}, XGBoost \cite{chen2016xgboost}, Random Forest \cite{breiman2001random}, CatBoost \cite{prokhorenkova2018catboost}, and LightGBM \cite{ke2017lightgbm} are preferred for their ability to handle complex non-linear relationships. These models are noted for their interpretability and effectiveness in dealing with diverse feature types, including missing values and categorical features.

Recent trends in the tabular data domain have seen the adoption of deep learning models designed to enhance performance. These include a variety of neural architectures such as ResNet \cite{he2016deep}, SNN \cite{klambauer2017self}, AutoInt \cite{song2019autoint}, and DCN V2 \cite{wang2021dcn}, which are primarily supervised methods. Additionally, hybrid approaches combine decision trees with neural networks for end-to-end training, examples of which include NODE \cite{popov2019neural}, GrowNet \cite{badirli2020gradient}, TabNN \cite{ke2018tabnn}, and DeepGBM \cite{ke2019deepgbm}. There are also emerging focuses on representation learning methods that utilize self- and semi-supervised learning for effective information extraction, such as VIME \cite{yoon2020vime}, SCARF \cite{bahri2021scarf}, and Recontab \cite{chen2023recontab}.

\subsection{Transformer-based Models}
In the realm of tabular data, transformer-based methods have become increasingly prominent, leveraging attention mechanisms to discern relationships across features and data samples. Notable models in this category include TabNet \cite{arik2021tabnet}, TabTransformer \cite{huang2020tabtransformer}, FT-Transformer \cite{gorishniy2021revisiting}, and SAINT \cite{somepalli2021saint}. These models exemplify the integration of transformer technology in tabular learning, offering enhanced modeling capabilities through attention-driven feature interactions.

\subsection{Contrastive Representation Learning}
Contrastive Representation Learning (CRL) has been substantively developed through significant contributions across various fields. The introduction of MoCo, a dynamic dictionary technique for unsupervised visual representation learning, marked a major advancement in the efficiency of learning algorithms \cite{he2020momentum}. Following this, the SimCLR framework simplified and improved the process by applying a straightforward contrastive loss to visual representations \cite{chen2020simple}. In the natural language processing arena, enhancements in sentence embeddings have been demonstrated through an efficient learning framework that applies CRL principles \cite{logeswaran2018efficient}. Furthermore, the application of CRL to supervised learning scenarios has shown substantial improvements in classifier robustness and accuracy, broadening the potential uses of this approach \cite{khosla2020supervised}. Recently, efforts inspired by CRL have been extended to the tabular domain. However, these initiatives primarily focus on instance-level contrast, which may limit their broader applicability \cite{chen2023recontab,somepalli2021saint,yoon2020vime}.

\subsection{Feature Decoupling}
Feature decoupling is integral to advancing machine learning models, enhancing both interpretability and performance by separating complex, intertwined data elements. In computer vision, techniques such as unsupervised domain-specific deblurring leverage disentangled representations to improve image clarity by effectively isolating content from blur features \cite{lu2019unsupervised}. The introduction of disentangled non-local neural networks demonstrates significant improvements in context modeling, benefiting tasks like semantic segmentation and object detection \cite{yin2020disentangled}. In multimodal transformers, decoupling strategies have been particularly effective, as demonstrated in zero-shot semantic segmentation, where the separation of components allows for better utilization of vision-language pre-trained models \cite{ding2022decoupling}. Moreover, in the tabular data domain, SwitchTab  \cite{wu2024switchtab} showcases feature decoupling to enhance self-supervised learning by effectively isolating mutual and salient features to improve decision-making and model robustness in downstream tasks. This work has inspired us to propose a more comprehensive contrastive learning framework to enhance the decoupled feature learning.

\section{Method}
In this section, we introduce TabDeco, our comprehensive framework for contrastive learning tailored to tabular data representation. We begin with outlining the supervised training process of TabDeco, setting the foundation for effective feature extraction, decoupling and contrastive learning approaches. We then introduce the core component of TabDeco, the global-local feature decoupling mechanisms, to enhance the model's adaptability and robustness across diverse datasets. Furthermore, we explore the integration of various contrastive loss combinations, illustrating how each tailored loss function uniquely contributes to improving the model's performance by optimizing representation learning with better separation and alignment in feature space. Finally, we summarize the supervised learning algorithm and discuss the training strategies.

\begin{figure*}[t!]
\begin{center}
    \includegraphics[width=1\linewidth]{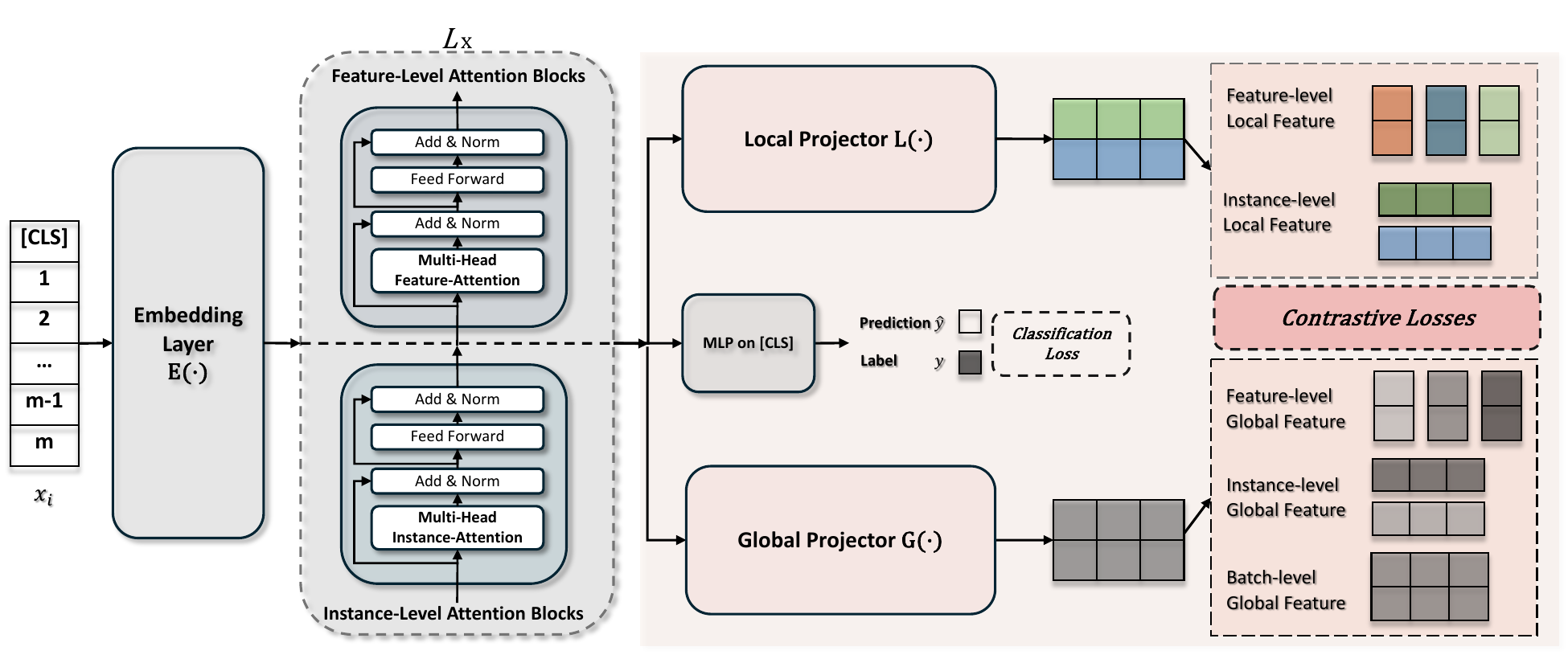}
\end{center}
  \caption{An overview of the TabDeco Framework.}
\label{fig: framework}
\end{figure*}

\subsection{Supervised Training Framework}
The supervised training architecture in our framework integrates column and row attention blocks with feature decoupling and contrastive learning, specifically tailored for tabular data. 
The attention blocks are from the state-of-the-art transformer-based model architectures considering both column-wise attentions for feature representation \cite{huang2020tabtransformer, gorishniy2021revisiting} and row-wise attentions for intersample representation \cite{somepalli2021saint}. The feature decoupling module is inspired by the breakthrough idea in SwitchTab \cite{wu2024switchtab}, further enhanced by the comprehensive contrastive learning process. Meanwhile, the natural integration of highly-resolved attention mechanism and fine-grained feature distinction could systematically enhance the model's ability to capture complex interactions among features and instances, making the most out of the unique characteristics of tabular data.

Given a tabular dataset represented by $\mathcal{D} = \{(\mathbf{x_i}, y_i)\}_{i=1}^N$, where each $\mathbf{x_i}$ is a $m\text{-dimensional}$ feature vector with $y_i$ as its associated label. $N$ is the total number of samples. As shown in Figure~\ref{fig: framework}, we append a $[\mathbf{CLS}]$ token with a learned embedding to each data sample, making $\mathbf{x_i}=[[\mathbf{CLS}], f_i^{1}, f_i^{2}, \cdots, f_i^{m}]$ be the single data point with categorical or numerical features. $\mathbf{E}$ is the embedding layer using different embedding functions to embed each feature into a $d\text{-dimensional}$ space, i.e., $\mathbf{x_i} \in \mathbb{R}^{(m+1)} \rightarrow E(\mathbf{x_i}) \in \mathbb{R}^{(m+1) \times d}$. The encoded representations are then passed through $L$-layer attention blocks composed of feature-level and instance-level attention mechanisms. Feature-level attention focuses on refining relationships among features (columns), while instance-level attention blocks handle the interactions among instances (rows). This dual attention mechanism ensures the model's adaptability to complex feature interactions and intersample relationships.

The outputs from these attention blocks are projected into local and global spaces via respective projectors: the Local Projector $\mathbf{L}(\cdot)$ generates a local feature vector to represent both feature-level and instance-level local features, whereas the Global Projector $\mathbf{G}(\cdot)$ produces a global feature vector which could simutaneously represent feature-level, instance-level, and batch-level global features. The local and global feature vectors are then used to compute various contrastive losses (e.g., InfoNCE) to enhance the selected distinctions among feature, instance and/or global patterns. Additionally, supervised loss (e.g., Cross Entropy), is integrated to directly guide the learning process towards label prediction. This structured approach allows the model to capture fine-grained feature relationships and improve predictive performance across different data distributions and complexities.

\subsection{Global-Local Decoupling}
We introduce the global-local decoupling mechanism to effectively capture information at multiple levels. Specifically, each feature vector $\mathbf{x_i}$ is decomposed into two distinct representations: a global feature vector $\mathbf{g_i}$ and a local feature vector $\mathbf{l_i}$. The global vector $\mathbf{g_i}$ captures broad patterns and shared characteristics across the entire dataset, focusing on the correlations among features and interrelationships across different samples. In contrast, the local vector $\mathbf{l_i}$ captures instance-specific variations that highlights the individual distinctions within each sample as well as the feature-specific uniqueness to distinguish one feature from the others. Mathematically, we define the global and local feature mappings as:
\begin{equation}
    \mathbf{g_i} = \mathbf{G}(\mathbf{x_i}; \theta_g), \quad \mathbf{l_i} = \mathbf{L}(\mathbf{x_i}; \theta_l)
\end{equation}
where $\mathbf{G}(\cdot)$ and $\mathbf{L}(\cdot)$ are learnable functions parameterized by $\theta_g$ and $\theta_l$ respectively.

\subsection{Contrastive Losses}
Given the decoupled global and local feature vectors $\mathbf{g_i}$ and $\mathbf{l_i}$, we are constructing positive and negative pairs to distinguish between broader feature distributions and fine-grained individual characteristics. In general, we are encouraging global features to be similar to other global features, and discouraging local features from being similar to other local features. In addition, global features are discouraged from being similar to local features. The loss functions are demoted as 
\begin{equation} \label{loss_global}
    \mathcal{L}_{\text{global}} = -\log \frac{\exp\left(\mathbf{sim}(\mathbf{g_i}, \mathbf{g_j})\\/{\tau}\right)}{\sum_{k} \exp\left(\mathbf{sim}(\mathbf{g_i}, \mathbf{g_k})\\/{\tau}\right)} , 
\end{equation}

\begin{equation} \label{loss_local}
    \mathcal{L}_{\text{local}} = -\log \frac{\exp\left(-\mathbf{sim}(\mathbf{l_i}, \mathbf{l_j})\\/{\tau}\right)}{\sum_{k} \exp\left(-\mathbf{sim}(\mathbf{l_i}, \mathbf{l_k})\\/{\tau}\right)} , 
\end{equation}

\begin{equation} \label{loss_cross}
    \mathcal{L}_{\text{cross}} = -\log \frac{\exp\left(-\mathbf{sim}(\mathbf{g_i}, \mathbf{l_j})\\/{\tau}\right)}{\sum_{k} \exp\left(-\mathbf{sim}(\mathbf{g_i}, \mathbf{l_k})\\/{\tau}\right)} , 
\end{equation}
where $\tau$ is the temperature parameter used to scale the similarity values and $k$ is the index across the entire set of features to compute the normalization factor for the similarity scores. $\mathbf{sim}(\cdot)$ represents the similarity measure between two vectors, e.g., cosine similarity.

Depending on the level of distinctiveness to be achieved, we proposed six types of contrastive losses, as shown in Table~\ref{Contrastive_losses}. Each loss function consists of the three loss components shown in Equation~(\ref{loss_global}-\ref{loss_cross}) with its corresponding similarity measure aggregating from the decoupled global and local feature vectors $\mathbf{g_i}$ and $\mathbf{l_i}$.
The major difference lies in the similarity measures with different summation logic to distinguish aspects of global and local features. Specifically, $\mathcal{L}_{\text{all}}$ aims to contrast each feature of each instance across the entire batch, creating a comprehensive similarity matrix of size $(b*m, b*m)$. It ensures that every data point is contrasted against all others, enforcing a comprehensive alignment that helps the model to learn subtle distinctions between different features and instances. Intuitively, this broad-level contrasting pushes the model to learn detailed, high-granularity representations by maximizing the separation between different feature-instance combinations. $\mathcal{L}_{\text{gg}}$ compares each batch of features to a shared, global feature representation randomly initialized with dimension $(m,d)$, producing a $(b,m,m)$ similarity structure. By contrasting batch-level features against a global standard, this loss encourages the model to align local features with global patterns, improving consistency across batches. The intuition here is to synchronize individual batches with a common global feature representation, promoting a unified feature understanding across diverse instances. 

\begin{figure*}[h]
\begin{center}
    \includegraphics[width=0.85\linewidth]{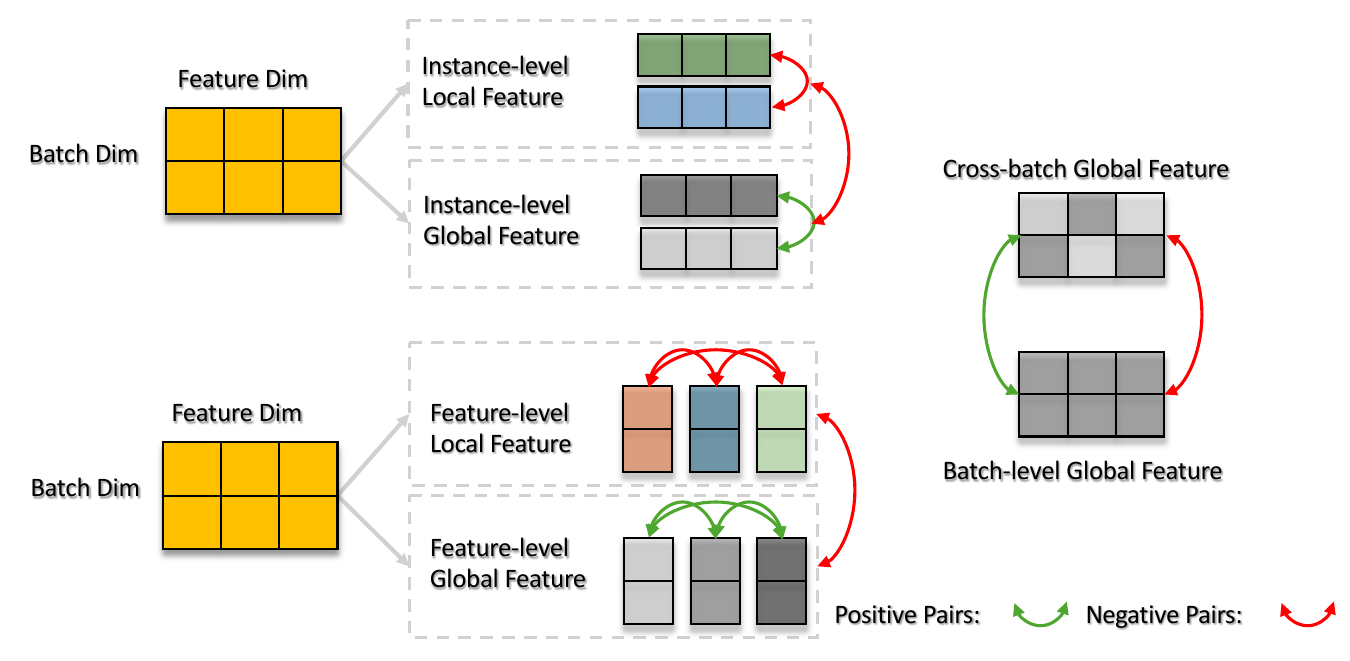}
\end{center}
  \caption{A demonstration of different levels of contrasts}
\label{fig: contrast}
\end{figure*}

The feature-level loss $\mathcal{L}_{\text{f}}$ contrasts each feature across all instances within the same batch and captures relationships between different features within the same batch, encouraging the model to differentiate features that frequently occur together while still learning their distinct roles. The intuition is to refine feature boundaries, making the model more adept at identifying individual feature influences within complex interactions. $\mathcal{L}_{\text{fs}}$ further contrasts each feature for each single instance across the batch and targets the fine-grained relationships of specific features of individual instances and fosters detailed understanding of how particular features vary between samples. Similarly, this sample-level loss $\mathcal{L}_{\text{s}}$ contrasts each instance by comparing all features within that instance against others in the batch. This approach focuses on enhancing the discriminative ability of the model at the instance level, making it better at distinguishing between different samples based on their overall feature profiles. The intuitive aim is to fine-tune how the model distinguishes between individual data points, aiding in more accurate instance-level classification. Furthermore, $\mathcal{L}_{\text{sf}}$ contrasts instances one feature at a time to emphasize on unique features with more differentiating power. The intuition is to enhance the model’s sensitivity to how each feature contributes to instance-level differences, sharpening the model's ability to make precise comparisons between instances. Algorithm~\ref{algo:DA} shows the details on how to integrate the global-local decoupling with contrastive learning through different types of losses and positive and negative pairs generation. In practices, different types of contrastive losses can be combined together to learn the feature and instance representations with different granularities. Figure~\ref{fig: contrast} further visualize the different levels of contrastive learning across global and local features. 

\begin{table*}[h]
    \centering
    
    \resizebox{1\textwidth}{!}{
    \begin{tabular}{lcl}
    \toprule
    \textbf{Loss} & \textbf{Similarity Summation} & \textbf{Description}\\
    \midrule
    $\mathcal{L}_{\text{all}}$          & $(b*m, d), (b*m, d) \rightarrow (b*m, b*m)$  & Contrasting each feature of each instance across the batch \\
    $\mathcal{L}_{\text{gg}}$          & $(b, m, d), (m, d) \rightarrow (b, m, m)$ & Contrasting each batch with the cross-batch feature-level global representation \\
    $\mathcal{L}_{\text{f}}$          & $(b, m, d), (b, m, d) \rightarrow (m, m)$ & Contrasting each feature of all instances in the batch\\
    $\mathcal{L}_{\text{s}}$         & $(b, m, d), (b, m, d) \rightarrow (b, b)$ & Contrasting each instance of all features in the batch\\
    $\mathcal{L}_{\text{fs}}$          & $(b, m, d), (b, m, d) \rightarrow (b, m, m)$ & Contrasting each feature for each instance across the batch\\
    $\mathcal{L}_{\text{sf}}$          & $(b, m, d), (b, m, d) \rightarrow (b, b, m)$ & Contrasting each instance across the batch for each feature\\
    \bottomrule
    \end{tabular}
    }
    \caption{Contrastive losses}
    \label{Contrastive_losses}
\end{table*}

\begin{algorithm}[h]
\caption{Supervised Learning with TabDeco}
\label{algo:DA}
\begin{algorithmic}[1]
{
\footnotesize
\REQUIRE training data with label $\mathcal{D} = \{(\mathbf{x_i}, y_i)\}_{i=1}^N$, number of features $m$, batch size $b$, embedding layer $\mathbf{E}$, embedding size $d$, attention blocks $\mathbf{A}$, projector for global feature $\mathbf{G}$, projector for local feature $\mathbf{L}$, multi-layer perceptron (MLP), cross-entropy (CE), mean squared error (MSE).

\FOR {one mini-batch $\left\{(\mathbf{x_i}, y_i)\right\}_{i=1}^{b} \subseteq \mathcal{D}$, $\mathbf{x_i} \in \mathbb{R}^{(m)}$} 
    \STATE for each sample $\mathbf{x_i}$, embed each feature into $d$-dimensional embeddings: 
    
    $\mathbf{\breve{x}_{i}} = \mathbf{E}(\mathbf{x_i}) \in \mathbb{R}^{m \times d}$
    
    \STATE for the batch of samples $\mathcal{X}^b = \left\{\mathbf{\breve{x}_{i}}\right\}_{i=1}^{b}$, embeddings are learned through the attention blocks:
    
    $\mathcal{Z}^b = \mathbf{A}(\mathcal{X}^b), \mathcal{Z}^b \in \mathbb{R}^{b \times m \times d}$
    
    \STATE Decouple the embedding sample into global and local vectors:  
    
    $\mathbf{g}^b = \mathbf{G}(\mathcal{Z}^b), \quad \mathbf{l}^b = \mathbf{L}(\mathcal{Z}^b), \quad \mathbf{g}^b, \mathbf{l}^b \in \mathbb{R}^{b \times m \times d}$
    
    \STATE Supervised learning prediction:
    
    $\hat{y}^b = \text{MLP}(\mathcal{Z}^b, \mathbf{g}^b, \mathbf{l}^b), \quad \text{where } \hat{y}^b = \left\{\hat{y}_i\right\}_{i=1}^{b} $

    \STATE Supervised loss computation:

    $\mathcal{L}_{\text{sup}} = \text{CE}(y_i, \hat{y}_i)$ for classification or $\mathcal{L}_{sup} = \text{MSE}(y_i, \hat{y}_i)$ for regression
    
    \STATE Contrastive loss computation: 
    
    (1) select the loss types, e.g., $\mathcal{L}_{\text{all}}$ or choose multiple ones $\mathcal{L}_{\text{f}}$, $\mathcal{L}_{\text{s}}$

    (2) for each loss type, compute the loss components from Equation~(\ref{loss_global}-\ref{loss_cross}), e.g., 

    $\mathcal{L}_{\text{all}} = \mathcal{L}_{\text{global}} + \mathcal{L}_{\text{local}} + \mathcal{L}_{\text{cross}}$

    (3) get the total contrastive loss

    $\mathcal{L}_{\text{contrast}} = \mathcal{L}_{\text{all}}$ or $\mathcal{L}_{\text{contrast}} = \mathcal{L}_{\text{f}} + \mathcal{L}_{\text{s}}$

    \STATE combine supervised loss with contrastive losses to get the total

    $\mathcal{L}_{\text{total}} = \mathcal{L}_{\text{sup}} + \alpha\mathcal{L}_{\text{contrastive}}$ where $\alpha$ is the weight balancing the contributions of each loss.

    \STATE update embedding layer $\mathbf{E}$, attention blocks $\mathbf{A}$, projectors $\mathbf{G}$ and $\mathbf{L}$, and minimize $\mathcal{L}_{total}$ through backpropagation.

\ENDFOR
}
\end{algorithmic}
\end{algorithm}

\section{Experiments and Results}
\subsection{Datasets, Setup, and Baselines}
\subsubsection{Datasets}
We evaluate the proposed method using a selection of widely recognized datasets, as utilized in recent studies \cite{somepalli2021saint}. These datasets include Bank (BK) \cite{moro2014data}, Blastchar (BC) \cite{ouktelco}, Shoppers (SH) \cite{sakar2019real}, Volkert (VO) \cite{automlchallenges} and MNIST (MN) \cite{xiao2017fashion}, among others. The datasets are diverse, ranging in size from 200 to 495,141 samples and spanning 8 to 784 features, incorporating both categorical and continuous variables. Some datasets contain missing data, while others are complete; similarly, some are well-balanced, whereas others exhibit highly skewed class distributions. All of these datasets are publicly accessible as shown in table~\ref{tab: source}.

\subsubsection{Model variants}
The TabDeco architecture discussed above follows similar design in SAINT to define three variants of model structures with different choices of attention block. TabDeco has the attention transformer encoder stacking both feature-level and instance-level attention blocks. TabDeco-s has only instance-level attention (e.g., SAINT-s) while TabDeco-f has only feature-level attention (e.g., SAINT-f).

\subsubsection{Training Details}
We train all models, including those with pre-training, using the AdamW optimizer with parameters set to \(\beta_1 = 0.9\), \(\beta_2 = 0.999\), a weight decay of 0.01, and a learning rate of 0.0001. The batch size is set to 128, except for datasets with a large number of features, such as MNIST and Volkert, where we use smaller batch sizes. The data is split into 65\% for training, 15\% for validation, and 20\% for testing. For the contrastive losses, we use temperature $\tau$ = 0.5. In each of our experiments, we use one NVIDIA A10G Tensor Core GPU. Individual training runs take between 5 minutes and 10 hours. For most of the datasets, we use embedding size $d$ = 32. Due to the memory constraints of a single GPU, we use $d$ = 4 for MNIST and $d$ = 8 for Volkert. We used $L$ = 6 layers and $h$ = 8 attention heads in all datasets except MNIST and Volkert, where we reduce to $L$ = 2 and $h$ = 4. We leverage the implementations of various methods in TALENT toolbox \cite{liu2024talent} for training and comparisons.

\subsubsection{Metrics}
Since most of the tasks in our analysis focus on binary classification, we primarily use the Area Under the Receiver Operating Characteristic curve (AUROC) to evaluate performance. AUROC provides a robust measure of the model's ability to differentiate between the two classes within the dataset. For the two multi-class datasets, Volkert (VO) and MNIST (MN), we instead use accuracy on the test set as the performance metric. This approach allows us to appropriately assess the model's effectiveness across different types of classification tasks.

\subsubsection{Baselines}
Following the previous paradigm \cite{somepalli2021saint,wu2024switchtab}, we conduct a comprehensive comparison of our model against a range of established techniques, including traditional algorithms like logistic regression and random forests, as well as advanced boosting frameworks such as XGBoost, LightGBM, and CatBoost. Additionally, we evaluate our model's performance alongside state-of-the-art deep learning models, including multi-layer perceptrons, VIME\cite{yoon2020vime}, TabNet\cite{arik2021tabnet}, TabTransformer\cite{wang2022transtab} and the three variants of SAINT \cite{somepalli2021saint}. SAINT by itself include both feature-level attention block and instance-level attention block in each stage. For the two variants, SAINT-f has only feature-level attention and SAINT-s has only intersample attention. For models that incorporate unsupervised pre-training, We focus on comparing the supervised performance and only considering the classification tasks.

\subsection{Main Results}

We report performance comparisons on the set of public datasets, with the AUROC results summarized in Table~\ref{tab: benchmark} for model predicting power. The corresponding standard errors are reported in Appendix in Table~\ref{tab: variance} for model robustness. Each number reported in the two tables represents the mean of 10 trials with different random seeds. 

In Table ~\ref{tab: benchmark}, TabDeco demonstrates significant predicting power ranking the best or second best across all datasets. Specifically, in 7 out of 11 datasets, one of the TabDeco variants outperforms all baseline models. In the remaining 4 datasets, TabDeco achieves the second best twice in Bank, outperformed by LightGBM and SAINT-f respectively. Income and Spambase datasets are also getting inferior performance from TabDeco, for which boosting methods have dominating predicting power. Meanwhile, TabDeco variants have shown the consistency to outperform their corresponding SAINT variants and SwitchTab for 10 datasets except Blastchar, which indicates the comprehensive enhancement from feature decoupling and constrastive learning. Table~\ref{tab: variance} further demonstrates the robustness of model performance for TabDeco, which exhibits better consistency than boosting methods and SAINT variants, especially for the datasets with better predicting power, such as Shoppers, HTRU2, etc. 

\begin{table*}[h]\renewcommand{\arraystretch}{0.9}\addtolength{\tabcolsep}{0pt}
    \centering
    
    \resizebox{1\textwidth}{!}{
    \begin{tabular}{lccccccccccc}
    \toprule
    \textbf{Dataset size}   & 45,211 & 7,043 & 12,330 & 32,561 & 17,898 & 10,000 & 4,601 & 1,055 & 495,141 & 58,310 & 60,000 \\   
    \textbf{Feature size}   & 16 & 27 & 17 & 14 & 8 & 11 & 57 & 41 & 49 & 147 & 784 \\   
    \midrule
    \textbf{Method/Dataset} & \textbf{Bank} & \textbf{Blastchar} & \textbf{Shoppers} & \textbf{Income} & \textbf{HTRU2} & \textbf{Shrutime} & \textbf{Spambase} & \textbf{QSARBio} & \textbf{Forest} & \textbf{Volkert$\dagger$} & \textbf{MNIST$\dagger$} \\
    \midrule
    Logistic Reg.     & 90.73 & 82.34 & 87.03 & 92.12 & 98.23 & 83.37 & 92.77 & 84.06 & 84.79 & 53.87 & 89.89 \\
    Random Forest     & 89.12 & 80.63 & 89.87 & 88.04 & 96.41 & 80.87 & 98.02 & 91.49 & 98.80 & 66.25 & 93.75 \\
    XGBoost           & 92.96 & 81.78 & 92.51 & \underline{92.31} & 97.81 & 83.59 & \underline{98.91} & 92.70 & 95.53 & 68.95 & 94.13 \\
    LightGBM          & \textbf{93.39} & 83.17 & 93.20 & \textbf{92.57} & 98.10 & 85.36 & \textbf{99.01} & 92.97 & 93.29 & 67.91 & 95.20 \\
    CatBoost          & 90.47 & 84.77 & 93.12 & 90.80 & 97.85 & 85.44 & 98.47 & 93.05 & 85.36 & 66.37 & 96.60 \\
    MLP               & 91.47 & 59.63 & 84.71 & 92.08 & 98.35 & 73.70 & 66.74 & 79.66 & 96.81 & 63.02 & 93.87 \\
    VIME              & 76.64 & 50.08 & 74.37 & 88.98 & 97.02 & 70.24 & 69.24 & 81.04 & 75.06 & 64.28 & 95.77 \\
    TabNet            & 91.76 & 79.61 & 91.38 & 90.72 & 97.58 & 75.24 & 97.93 & 67.55 & 96.37 & 56.83 & 96.79 \\
    TabTransormer     & 91.34 & 81.67 & 92.70 & 90.60 & 96.56 & 85.60 & 98.50 & 91.80 & 84.96 & 57.98 & 88.74  \\
    SwitchTab         & 91.80 & {83.56} & 91.20 & 89.79 & 97.15 & 84.89 & 96.06 & \underline{93.85} & 97.66 & 68.90 & 96.80  \\
    \midrule
    SAINT             & {93.12} & 84.58 & \underline{93.22} & {90.99} & 98.28 & {85.79} & {97.66} & 93.34 & 99.06 & 68.87 & {97.48} \\
    SAINT-s           & {92.84} & 83.57 & {92.31} & {90.59} & 98.58 & {85.43} & {97.83} & 92.48 & 98.85 & \underline{69.61} & {97.45} \\
    SAINT-f           & {92.68} & \textbf{85.18} & {93.10} & 91.23 & 98.54 & {85.09} & {97.40} & 91.91 & 98.96 & 58.76 & {91.25} \\
    \midrule
    TabDeco           & \underline{93.34} & 84.96 & \textbf{93.53} & {91.17} & 98.57 & \textbf{86.14} & {97.79} & {93.80} & \textbf{99.15} & {69.46} & \textbf{97.85} \\
    TabDeco-s         & {93.00} & 84.16 & {92.66} & {90.95} & \textbf{98.60} & \underline{85.93} & {97.97} & \textbf{94.87} & \underline{99.06} & \textbf{69.91} & \underline{97.71} \\
    TabDeco-f         & {92.48} & \underline{85.00} & {93.05} & {91.26} & \textbf{98.60} & {85.00} & {97.36} & 92.08 & 99.00 & 58.52 & {95.56} \\
    \bottomrule
    \end{tabular}
    }
    \caption{\small Mean AUROC scores averaged over 10 trials for 11 datasets on classification tasks. Baseline results are quoted from original papers when possible and reproduced otherwise. We highlight best result in bold and second best in underline. Columns denoted by $\dagger$ are multi-class problems, and we report accuracy.}
    \label{tab: benchmark}
\end{table*}

\subsection{Ablation Studies}

In this section, we conduct experiments on different contrastive loss combinations to evaluate how they could performance differently for various datasets. We test adding the loss combination to the total loss for each TabDeco variant and report the best performance for each combination from all variants. The baseline noted by ``-'' is without any contrastive losses or feature decoupling, deteriorating to the SAINT variants for classification-only tasks. To determine whether the loss combinations are useful, we count the ones outperforming baseline for each dataset. Similarly, we count the outperforming ones for each loss combination across the datasets to check the consistency. As shown in Table`\ref{tab: ablation_losses}, out of the 11 datasets, losses for more comprehensive feature-level (e.g., $\mathcal{L}_{\text{fs}}$) and/or instance-level (e.g., $\mathcal{L}_{\text{sf}}$) and/or cross-batch contrasting (e.g., $\mathcal{L}_{\text{gg}}$) and their combinations are demonstrated to be significantly superior on getting better performance, outperforming the baseline for at least 6 and up to 11 datasets. For each datasets, the performance may vary across different contrastive losses. Out of the 14 combinations we test, the counts that outperform baseline varies from 1 to 10. Specifically the simplified feature-level or instance-level losses $\mathcal{L}_{\text{f}}$ and  $\mathcal{L}_{\text{s}}$ can hardly get better performance. The global contrastive loss $\mathcal{L}_{\text{gg}}$ as well as the one capturing more granular level feature and/or instance characteristics generally perform better.

\begin{table*}[h]
    \centering
    
    \resizebox{1\textwidth}{!}{
    \begin{tabular}{lccccccccccccc|c}
    \toprule
    \textbf{Dataset/Loss} & \textbf{--} &  $\mathcal{L}_{\text{f}}$ &  $\mathcal{L}_{\text{s}}$ &  $\mathcal{L}_{\text{f+s}}$ &  $\mathcal{L}_{\text{fs}}$ &  $\mathcal{L}_{\text{sf}}$ &  $\mathcal{L}_{\text{fs+sf}}$ &  $\mathcal{L}_{\text{all}}$ &  $\mathcal{L}_{\text{gg}}$ &  $\mathcal{L}_{\text{fs+gg}}$ &  $\mathcal{L}_{\text{sf+gg}}$ &  $\mathcal{L}_{\text{all+gg}}$ &  $\mathcal{L}_{\text{sf+fs+gg}}$ & Counts by Row (14)\\
    \midrule
    Bank          & 93.12 & 92.56 & 93.08 & 92.69 & 93.29 & 93.21 & 93.22 & 93.20 & 93.13 & 93.34 & 93.23 & 92.99 & 93.15 & 8 \\
    Blastchar     & 84.58 & 82.61 & 84.56 & 84.51 & 84.75 & 84.96 & 84.51 & 84.79 & 84.86 & 85.00 & 84.71 & 84.96 & 85.00 & 8 \\
    Shoppers      & 93.22 & 90.19 & 92.95 & 92.18 & 93.32 & 93.15 & 93.24 & 93.29 & 93.24 & 93.29 & 93.09 & 93.06 & 93.53 & 6 \\
    Income        & 91.23 & 91.00 & 90.70 & 90.73 & 91.15 & 91.19 & 91.05 & 91.14 & 91.18 & 91.16 & 91.20 & 91.17 & 91.26 & 1 \\
    HTRU2         & 98.58 & 98.07 & 98.59 & 98.21 & 98.60 & 98.56 & 98.57 & 98.49 & 98.57 & 98.58 & 98.56 & 98.46 & 98.60 & 5 \\
    Shrutime      & 85.79 & 78.32 & 85.45 & 77.19 & 85.75 & 86.14 & 85.67 & 85.84 & 86.08 & 85.79 & 85.94 & 86.03 & 85.82 & 7 \\
    Spambase      & 97.83 & 93.00 & 97.97 & 91.96 & 97.77 & 97.96 & 97.94 & 97.83 & 97.83 & 97.82 & 97.91 & 97.89 & 97.93 & 7 \\
    QSARBio       & 93.34 & 84.27 & 94.87 & 83.14 & 94.33 & 93.79 & 94.50 & 94.34 & 93.45 & 94.13 & 93.71 & 94.27 & 93.80 & 10 \\
    Forest        & 99.06 & 94.35 & 97.89 & 97.50 & 98.99 & 99.15 & 99.10 & 96.18 & 99.06 & 98.96 & 99.10 & 98.26 & 99.08 & 5 \\
    Volkert$\dagger$       & {69.61} & 62.27 & {64.86} & {61.28} & 69.88 & {69.50} & {69.70} & 69.91 & 69.58 & 69.78 & 69.88 & 69.58 & 69.67 & 6 \\
    MNIST$\dagger$        & {97.48} & 91.02 & {93.96} & {63.58} & 97.39 & {97.13} & {97.58} & 96.96 & 97.58 & 97.44 & 97.85 & 97.37 & 97.83 & 4 \\
    \midrule
    Counts by Column (11)       & -- & 0 & {3} & {0} & 6 & {6} & {6} & 7 & 8 & 7 & 8 & 4 & 11\\
    \bottomrule
    \end{tabular}
    }
    \caption{\small Comparison of the best performance for each contrastive loss combination across TabDeco variants. ``-'' corresponds to the SAINT variants without feature decoupling or contrastive learning. We count the results by rows across datasets or by columns across lossess as long as it is better than ``-''. Rows denoted by $\dagger$ are multi-class problems, and we report accuracy rather than AUROC.}
    \label{tab: ablation_losses}
\end{table*}

\section{Conclusion}
In conclusion, the complexities inherent in tabular data demand innovative approaches that go beyond traditional deep learning and tree-based models. While methods like SwitchTab and SAINT have made strides in enhancing representation learning through feature decoupling and attention mechanisms, they fall short in fully addressing the challenges of dataset complexity and the generation of meaningful sample pairs. Our proposed framework, TabDeco, synthesizes the strengths of these methods by integrating attention-based contrastive learning with feature decoupling, enabling a deeper exploration of local and global interactions within the data. Extensive experiments demonstrate that TabDeco consistently outperforms existing models, including leading gradient boosting algorithms, across various benchmark tasks, underscoring its effectiveness and adaptability. By advancing the construction of positive and negative samples through multi-perspective contrastive learning, TabDeco sets a new standard for tabular data representation, offering a robust and interpretable solution for complex tabular scenarios. This work not only addresses current limitations but also paves the way for future innovations in contrastive learning for tabular data.

\bibliographystyle{plain}
\bibliography{neurips_2024}

\begin{thebibliography}{10}

\bibitem{arik2021tabnet}
Sercan~{\"O} Arik and Tomas Pfister.
\newblock Tabnet: Attentive interpretable tabular learning.
\newblock In {\em Proceedings of the AAAI conference on artificial intelligence}, volume~35, pages 6679--6687, 2021.

\bibitem{badirli2020gradient}
Sarkhan Badirli, Xuanqing Liu, Zhengming Xing, Avradeep Bhowmik, Khoa Doan, and Sathiya~S Keerthi.
\newblock Gradient boosting neural networks: Grownet.
\newblock {\em arXiv preprint arXiv:2002.07971}, 2020.

\bibitem{bahri2021scarf}
Dara Bahri, Heinrich Jiang, Yi~Tay, and Donald Metzler.
\newblock Scarf: Self-supervised contrastive learning using random feature corruption.
\newblock {\em arXiv preprint arXiv:2106.15147}, 2021.

\bibitem{borisov2022deep}
Vadim Borisov, Tobias Leemann, Kathrin Se{\ss}ler, Johannes Haug, Martin Pawelczyk, and Gjergji Kasneci.
\newblock Deep neural networks and tabular data: A survey.
\newblock {\em IEEE Transactions on Neural Networks and Learning Systems}, 2022.

\bibitem{breiman2001random}
Leo Breiman.
\newblock Random forests.
\newblock {\em Machine learning}, 45:5--32, 2001.

\bibitem{breiman2017classification}
Leo Breiman.
\newblock {\em Classification and regression trees}.
\newblock Routledge, 2017.

\bibitem{chen2020some}
Suiyao Chen.
\newblock Some recent advances in design of bayesian binomial reliability demonstration tests.
\newblock {\em USF Tampa Graduate Theses and Dissertations}, 2020.

\bibitem{chen2017personalized}
Suiyao Chen, William~D Kearns, James~L Fozard, and Mingyang Li.
\newblock Personalized fall risk assessment for long-term care services improvement.
\newblock In {\em 2017 Annual Reliability and Maintainability Symposium (RAMS)}, pages 1--7. IEEE, 2017.

\bibitem{chen2019claims}
Suiyao Chen, Nan Kong, Xuxue Sun, Hongdao Meng, and Mingyang Li.
\newblock Claims data-driven modeling of hospital time-to-readmission risk with latent heterogeneity.
\newblock {\em Health care management science}, 22:156--179, 2019.

\bibitem{chen2024deep}
Suiyao Chen, Xinyi Liu, Yulei Li, Jing Wu, and Handong Yao.
\newblock Deep representation learning for multi-functional degradation modeling of community-dwelling aging population.
\newblock {\em arXiv preprint arXiv:2404.05613}, 2024.

\bibitem{chen2017multi}
Suiyao Chen, Lu~Lu, and Mingyang Li.
\newblock Multi-state reliability demonstration tests.
\newblock {\em Quality Engineering}, 29(3):431--445, 2017.

\bibitem{chen2018data}
Suiyao Chen, Lu~Lu, Yisha Xiang, Qing Lu, and Mingyang Li.
\newblock A data heterogeneity modeling and quantification approach for field pre-assessment of chloride-induced corrosion in aging infrastructures.
\newblock {\em Reliability Engineering \& System Safety}, 171:123--135, 2018.

\bibitem{chen2023recontab}
Suiyao Chen, Jing Wu, Naira Hovakimyan, and Handong Yao.
\newblock Recontab: Regularized contrastive representation learning for tabular data.
\newblock {\em arXiv preprint arXiv:2310.18541}, 2023.

\bibitem{chen2016xgboost}
Tianqi Chen and Carlos Guestrin.
\newblock Xgboost: A scalable tree boosting system.
\newblock In {\em Proceedings of the 22nd acm sigkdd international conference on knowledge discovery and data mining}, pages 785--794, 2016.

\bibitem{chen2020simple}
Ting Chen, Simon Kornblith, Mohammad Norouzi, and Geoffrey Hinton.
\newblock A simple framework for contrastive learning of visual representations.
\newblock In {\em International conference on machine learning}, pages 1597--1607. PMLR, 2020.

\bibitem{ding2022decoupling}
Jian Ding, Nan Xue, Gui-Song Xia, and Dengxin Dai.
\newblock Decoupling zero-shot semantic segmentation.
\newblock In {\em Proceedings of the IEEE/CVF Conference on Computer Vision and Pattern Recognition}, pages 11583--11592, 2022.

\bibitem{gorishniy2021revisiting}
Yury Gorishniy, Ivan Rubachev, Valentin Khrulkov, and Artem Babenko.
\newblock Revisiting deep learning models for tabular data.
\newblock {\em Advances in Neural Information Processing Systems}, 34:18932--18943, 2021.

\bibitem{automlchallenges}
Isabelle Guyon, Lisheng Sun-Hosoya, Marc Boull\'e, Hugo~Jair Escalante, Sergio Escalera, Zhengying Liu, Damir Jajetic, Bisakha Ray, Mehreen Saeed, Mich\'ele Sebag, Alexander Statnikov, WeiWei Tu, and Evelyne Viegas.
\newblock Analysis of the automl challenge series 2015-2018.
\newblock In {\em AutoML}, Springer series on Challenges in Machine Learning, 2019.

\bibitem{hastie2017generalized}
Trevor~J Hastie and Daryl Pregibon.
\newblock Generalized linear models.
\newblock In {\em Statistical models in S}, pages 195--247. Routledge, 2017.

\bibitem{he2020momentum}
Kaiming He, Haoqi Fan, Yuxin Wu, Saining Xie, and Ross Girshick.
\newblock Momentum contrast for unsupervised visual representation learning.
\newblock In {\em Proceedings of the IEEE/CVF conference on computer vision and pattern recognition}, pages 9729--9738, 2020.

\bibitem{he2016deep}
Kaiming He, Xiangyu Zhang, Shaoqing Ren, and Jian Sun.
\newblock Deep residual learning for image recognition.
\newblock In {\em Proceedings of the IEEE conference on computer vision and pattern recognition}, pages 770--778, 2016.

\bibitem{huang2020tabtransformer}
Xin Huang, Ashish Khetan, Milan Cvitkovic, and Zohar Karnin.
\newblock Tabtransformer: Tabular data modeling using contextual embeddings.
\newblock {\em arXiv preprint arXiv:2012.06678}, 2020.

\bibitem{ke2017lightgbm}
Guolin Ke, Qi~Meng, Thomas Finley, Taifeng Wang, Wei Chen, Weidong Ma, Qiwei Ye, and Tie-Yan Liu.
\newblock Lightgbm: A highly efficient gradient boosting decision tree.
\newblock {\em Advances in neural information processing systems}, 30, 2017.

\bibitem{ke2019deepgbm}
Guolin Ke, Zhenhui Xu, Jia Zhang, Jiang Bian, and Tie-Yan Liu.
\newblock Deepgbm: A deep learning framework distilled by gbdt for online prediction tasks.
\newblock In {\em Proceedings of the 25th ACM SIGKDD International Conference on Knowledge Discovery \& Data Mining}, pages 384--394, 2019.

\bibitem{ke2018tabnn}
Guolin Ke, Jia Zhang, Zhenhui Xu, Jiang Bian, and Tie-Yan Liu.
\newblock Tabnn: A universal neural network solution for tabular data.
\newblock 2018.

\bibitem{khosla2020supervised}
Prannay Khosla, Piotr Teterwak, Chen Wang, Abhinav Sarna, Yonglong Tian, Phillip Isola, Aaron Maschinot, Ce~Liu, and Dilip Krishnan.
\newblock Supervised contrastive learning.
\newblock {\em Advances in Neural Information Processing Systems}, 33:18661--18673, 2020.

\bibitem{klambauer2017self}
G{\"u}nter Klambauer, Thomas Unterthiner, Andreas Mayr, and Sepp Hochreiter.
\newblock Self-normalizing neural networks.
\newblock {\em Advances in neural information processing systems}, 30, 2017.

\bibitem{lahn2023combinatorial}
Nathaniel Lahn, Sharath Raghvendra, and Kaiyi Zhang.
\newblock A combinatorial algorithm for approximating the optimal transport in the parallel and mpc settings.
\newblock {\em Advances in Neural Information Processing Systems}, 36:21675--21686, 2023.

\bibitem{lai2024residual}
Zhixin Lai, Jing Wu, Suiyao Chen, Yucheng Zhou, and Naira Hovakimyan.
\newblock Residual-based language models are free boosters for biomedical imaging tasks.
\newblock In {\em Proceedings of the IEEE/CVF Conference on Computer Vision and Pattern Recognition}, pages 5086--5096, 2024.

\bibitem{lai2024adaptive}
Zhixin Lai, Xuesheng Zhang, and Suiyao Chen.
\newblock Adaptive ensembles of fine-tuned transformers for llm-generated text detection.
\newblock {\em arXiv preprint arXiv:2403.13335}, 2024.

\bibitem{liakos2018machine}
Konstantinos~G Liakos, Patrizia Busato, Dimitrios Moshou, Simon Pearson, and Dionysis Bochtis.
\newblock Machine learning in agriculture: A review.
\newblock {\em Sensors}, 18(8):2674, 2018.

\bibitem{liu2024talent}
Si-Yang Liu, Hao-Run Cai, Qi-Le Zhou, and Han-Jia Ye.
\newblock Talent: A tabular analytics and learning toolbox.
\newblock {\em arXiv preprint arXiv:2407.04057}, 2024.

\bibitem{logeswaran2018efficient}
Lajanugen Logeswaran and Honglak Lee.
\newblock An efficient framework for learning sentence representations.
\newblock In {\em International Conference on Learning Representations}, 2018.

\bibitem{lu2019unsupervised}
Boyu Lu, Jun-Cheng Chen, and Rama Chellappa.
\newblock Unsupervised domain-specific deblurring via disentangled representations.
\newblock In {\em Proceedings of the IEEE/CVF Conference on Computer Vision and Pattern Recognition}, pages 10225--10234, 2019.

\bibitem{moro2014data}
S{\'e}rgio Moro, Paulo Cortez, and Paulo Rita.
\newblock A data-driven approach to predict the success of bank telemarketing.
\newblock {\em Decision Support Systems}, 62:22--31, 2014.

\bibitem{ouktelco}
James Ouk, David Dada, and Kyung~Tae Kang.
\newblock Telco customer churn.
\newblock 2018.

\bibitem{phatak2023computing}
Abhijeet Phatak, Sharath Raghvendra, Chittaranjan Tripathy, and Kaiyi Zhang.
\newblock Computing all optimal partial transports.
\newblock In {\em International Conference on Learning Representations}, 2023.

\bibitem{popov2019neural}
Sergei Popov, Stanislav Morozov, and Artem Babenko.
\newblock Neural oblivious decision ensembles for deep learning on tabular data.
\newblock {\em arXiv preprint arXiv:1909.06312}, 2019.

\bibitem{prokhorenkova2018catboost}
Liudmila Prokhorenkova, Gleb Gusev, Aleksandr Vorobev, Anna~Veronika Dorogush, and Andrey Gulin.
\newblock Catboost: unbiased boosting with categorical features.
\newblock {\em Advances in neural information processing systems}, 31, 2018.

\bibitem{qayyum2020secure}
Adnan Qayyum, Junaid Qadir, Muhammad Bilal, and Ala Al-Fuqaha.
\newblock Secure and robust machine learning for healthcare: A survey.
\newblock {\em IEEE Reviews in Biomedical Engineering}, 14:156--180, 2020.

\bibitem{raghvendra2024new}
Sharath Raghvendra, Pouyan Shirzadian, and Kaiyi Zhang.
\newblock A new robust partial $ p $-wasserstein-based metric for comparing distributions.
\newblock In {\em International Conference on Machine Learning}, 2024.

\bibitem{sakar2019real}
C~Okan Sakar, S~Olcay Polat, Mete Katircioglu, and Yomi Kastro.
\newblock Real-time prediction of online shoppers’ purchasing intention using multilayer perceptron and lstm recurrent neural networks.
\newblock {\em Neural Computing and Applications}, 31:6893--6908, 2019.

\bibitem{somepalli2021saint}
Gowthami Somepalli, Micah Goldblum, Avi Schwarzschild, C~Bayan Bruss, and Tom Goldstein.
\newblock Saint: Improved neural networks for tabular data via row attention and contrastive pre-training.
\newblock {\em arXiv preprint arXiv:2106.01342}, 2021.

\bibitem{song2019autoint}
Weiping Song, Chence Shi, Zhiping Xiao, Zhijian Duan, Yewen Xu, Ming Zhang, and Jian Tang.
\newblock Autoint: Automatic feature interaction learning via self-attentive neural networks.
\newblock In {\em Proceedings of the 28th ACM international conference on information and knowledge management}, pages 1161--1170, 2019.

\bibitem{tao2022optimizing}
Ran Tao, Pan Zhao, Jing Wu, Nicolas~F Martin, Matthew~T Harrison, Carla Ferreira, Zahra Kalantari, and Naira Hovakimyan.
\newblock Optimizing crop management with reinforcement learning and imitation learning.
\newblock {\em arXiv preprint arXiv:2209.09991}, 2022.

\bibitem{bingjie2023optimal}
Bingjie Wang, Lu~Lu, Suiyao Chen, and Mingyang Li.
\newblock Optimal test design for reliability demonstration under multi-stage acceptance uncertainties.
\newblock {\em Quality Engineering}, 0(0):1--14, 2023.

\bibitem{wang2021dcn}
Ruoxi Wang, Rakesh Shivanna, Derek Cheng, Sagar Jain, Dong Lin, Lichan Hong, and Ed~Chi.
\newblock Dcn v2: Improved deep \& cross network and practical lessons for web-scale learning to rank systems.
\newblock In {\em Proceedings of the web conference 2021}, pages 1785--1797, 2021.

\bibitem{wang2023balanced}
Yite Wang, Jing Wu, Naira Hovakimyan, and Ruoyu Sun.
\newblock Balanced training for sparse gans.
\newblock In {\em Thirty-seventh Conference on Neural Information Processing Systems}, 2023.

\bibitem{wang2022transtab}
Zifeng Wang and Jimeng Sun.
\newblock Transtab: Learning transferable tabular transformers across tables.
\newblock {\em Advances in Neural Information Processing Systems}, 35:2902--2915, 2022.

\bibitem{wright1995logistic}
Raymond~E Wright.
\newblock Logistic regression.
\newblock 1995.

\bibitem{wu2024switchtab}
Jing Wu, Suiyao Chen, Qi~Zhao, Renat Sergazinov, Chen Li, Shengjie Liu, Chongchao Zhao, Tianpei Xie, Hanqing Guo, Cheng Ji, et~al.
\newblock Switchtab: Switched autoencoders are effective tabular learners.
\newblock In {\em Proceedings of the AAAI Conference on Artificial Intelligence}, volume~38, pages 15924--15933, 2024.

\bibitem{wu2023hallucination}
Jing Wu, Jennifer Hobbs, and Naira Hovakimyan.
\newblock Hallucination improves the performance of unsupervised visual representation learning.
\newblock In {\em Proceedings of the IEEE/CVF International Conference on Computer Vision}, pages 16132--16143, 2023.

\bibitem{wu2023genco}
Jing Wu, Naira Hovakimyan, and Jennifer Hobbs.
\newblock Genco: An auxiliary generator from contrastive learning for enhanced few-shot learning in remote sensing.
\newblock {\em arXiv preprint arXiv:2307.14612}, 2023.

\bibitem{wu2024new}
Jing Wu, Zhixin Lai, Suiyao Chen, Ran Tao, Pan Zhao, and Naira Hovakimyan.
\newblock The new agronomists: Language models are experts in crop management.
\newblock In {\em Proceedings of the IEEE/CVF Conference on Computer Vision and Pattern Recognition}, pages 5346--5356, 2024.

\bibitem{wu2024crops}
Jing Wu, Zhixin Lai, Shengjie Liu, Suiyao Chen, Ran Tao, Pan Zhao, Chuyuan Tao, Yikun Cheng, and Naira Hovakimyan.
\newblock Crops: A deployable crop management system over all possible state availabilities.
\newblock {\em arXiv preprint arXiv:2411.06034}, 2024.

\bibitem{wu2023extended}
Jing Wu, David Pichler, Daniel Marley, David Wilson, Naira Hovakimyan, and Jennifer Hobbs.
\newblock Extended agriculture-vision: An extension of a large aerial image dataset for agricultural pattern analysis.
\newblock {\em arXiv preprint arXiv:2303.02460}, 2023.

\bibitem{wu2022optimizing}
Jing Wu, Ran Tao, Pan Zhao, Nicolas~F Martin, and Naira Hovakimyan.
\newblock Optimizing nitrogen management with deep reinforcement learning and crop simulations.
\newblock In {\em Proceedings of the IEEE/CVF conference on computer vision and pattern recognition}, pages 1712--1720, 2022.

\bibitem{xiao2017fashion}
Han Xiao, Kashif Rasul, and Roland Vollgraf.
\newblock Fashion-mnist: a novel image dataset for benchmarking machine learning algorithms.
\newblock {\em arXiv preprint arXiv:1708.07747}, 2017.

\bibitem{ye2024multiplexed}
Jiachi Ye, Haoyan Kang, Qian Cai, Zibo Hu, Maria Solyanik-Gorgone, Hao Wang, Elham Heidari, Chandraman Patil, Mohammad-Ali Miri, Navid Asadizanjani, et~al.
\newblock Multiplexed orbital angular momentum beams demultiplexing using hybrid optical-electronic convolutional neural network.
\newblock {\em Nature Communications Physics}, 7(1):105, 2024.

\bibitem{ye2023multiplexed}
Jiachi Ye, Haoyan Kang, Hao Wang, Salem Altaleb, Elham Heidari, Navid Asadizanjani, Volker~J Sorger, and Hamed Dalir.
\newblock Multiplexed oam beams classification via fourier optical convolutional neural network.
\newblock In {\em 2023 IEEE Photonics Conference (IPC)}, pages 1--2. IEEE, 2023.

\bibitem{ye2023oam}
Jiachi Ye, Haoyan Kang, Hao Wang, Salem Altaleb, Elham Heidari, Navid Asadizanjani, Volker~J Sorger, and Hamed Dalir.
\newblock Oam beams multiplexing and classification under atmospheric turbulence via fourier convolutional neural network.
\newblock In {\em Frontiers in Optics}, pages JTu4A--73. Optica Publishing Group, 2023.

\bibitem{ye2023demultiplexing}
Jiachi Ye, Haoyan Kang, Hao Wang, Chen Shen, Belal Jahannia, Elham Heidari, Navid Asadizanjani, Mohammad-Ali Miri, Volker~J Sorger, and Hamed Dalir.
\newblock Demultiplexing oam beams via fourier optical convolutional neural network.
\newblock In {\em Laser Beam Shaping XXIII}, volume 12667, pages 16--33. SPIE, 2023.

\bibitem{ye2023free}
Jiachi Ye, Maria Solyanik, Zibo Hu, Hamed Dalir, Behrouz~Movahhed Nouri, and Volker~J Sorger.
\newblock Free-space optical multiplexed orbital angular momentum beam identification system using fourier optical convolutional layer based on 4f system.
\newblock In {\em Complex Light and Optical Forces XVII}, volume 12436, pages 69--79. SPIE, 2023.

\bibitem{yin2020disentangled}
Minghao Yin, Zhuliang Yao, Yue Cao, Xiu Li, Zheng Zhang, Stephen Lin, and Han Hu.
\newblock Disentangled non-local neural networks.
\newblock In {\em Computer Vision--ECCV 2020: 16th European Conference, Glasgow, UK, August 23--28, 2020, Proceedings, Part XV 16}, pages 191--207. Springer, 2020.

\bibitem{yoon2020vime}
Jinsung Yoon, Yao Zhang, James Jordon, and Mihaela van~der Schaar.
\newblock Vime: Extending the success of self-and semi-supervised learning to tabular domain.
\newblock {\em Advances in Neural Information Processing Systems}, 33:11033--11043, 2020.

\end{thebibliography}


\appendix

\section{Appendix / supplemental material}

\subsection{Datasets Details}

\begin{table*}[h]
    \centering
    
    \resizebox{1\textwidth}{!}{
    \begin{tabular}{ll}
    \toprule
    \textbf{Dataset} & \textbf{Source Link}\\
    \midrule
    Bank          & \url{https://archive.ics.uci.edu/ml/datasets/bank+marketing}\\
    Blastchar     & \url{https://www.kaggle.com/blastchar/telco-customer-churn}\\
    Shoppers      & \url{https://archive.ics.uci.edu/ml/datasets/Online+Shoppers+Purchasing+Intention+Dataset}\\
    Income        & \url{https://www.kaggle.com/lodetomasi1995/income-classification}\\
    HTRU2         & \url{https://archive.ics.uci.edu/ml/datasets/HTRU2}\\
    Shrutime      & \url{https://www.kaggle.com/shrutimechlearn/churn-modelling}\\
    Spambase      & \url{https://archive.ics.uci.edu/ml/datasets/Spambase}\\
    QSARBio       & \url{https://archive.ics.uci.edu/ml/datasets/QSAR+biodegradation}\\
    Forest        & \url{https://kdd.ics.uci.edu/databases/covertype}\\
    Volkert       & \url{http://automl.chalearn.org/data}\\
    MNIST         & \url{http://yann.lecun.com/exdb/mnist/}\\
    \bottomrule
    \end{tabular}
    }
    \caption{Dataset links}
    \label{tab: source}
\end{table*}

\subsection{Additional Results}
\begin{table*}[h]
    \centering
    
    \resizebox{1\textwidth}{!}{
    \begin{tabular}{lccccccccccc}
    \toprule
    \textbf{Method/Dataset} & \textbf{Bank} & \textbf{Blastchar} & \textbf{Shoppers} & \textbf{Income} & \textbf{HTRU2} & \textbf{Shrutime} & \textbf{Spambase} & \textbf{QSARBio} & \textbf{Forest} & \textbf{Volkert$\dagger$} & \textbf{MNIST$\dagger$} \\
    \midrule
    Logistic Reg.     & 0.25 & 0.20 & 0.41 & 6.34 & 0.26 & 0.53 & 0.12 & 0.70 & 0.11 & 1.33 & 3.19 \\
    Random Forest     & 0.27 & 0.70 & 0.60 & 0.30 & 0.25 & 0.38 & 0.27 & 0.80 & 0.01 & 1.27 & 4.59 \\
    XGBoost           & 0.15 & 0.34 & 0.50 & 0.15 & 0.10 & 0.39 & 0.08 & 0.45 & 0.01 & 0.51 & 1.98 \\
    LightGBM          & 0.21 & 0.34 & 0.48 & 0.13 & 0.13 & 0.58 & 0.05 & 0.67 & 0.01 & 0.64 & 3.78 \\
    CatBoost          & 0.17 & 0.19 & 0.41 & 0.15 & 0.23 & 0.41 & 0.11 & 0.79 & 0.01 & 1.17 & 1.66 \\
    MLP               & 0.21 & 0.32 & 0.60 & 2.74 & 0.31 & 1.65 & 0.15 & 1.00 & 0.68 & 1.56 & 3.74 \\
    VIME              & 2.03 & 0.26 & 2.74 & 5.10 & 2.52 & 1.15 & 3.03 & 0.71 & 6.91 & 6.67 & 8.15 \\
    TabNet            & 0.33 & 0.30 & 0.68 & 0.17 & 0.29 & 5.12 & 0.15 & 2.67 & 0.01 & 1.47 & 2.22 \\
    TabTransormer     & 0.34 & 0.30 & 0.69 & 0.17 & 0.29 & 5.18 & 0.15 & 2.70 & 0.01 & 1.48 & 2.24  \\
    SwitchTab         & 0.54 & 0.39 & 0.65 & 0.47 & 0.45 & 0.76 & 0.44 & 1.55 & 0.03 & 0.38 & 0.54  \\
    \midrule
    SAINT             & {0.12} & 0.35 & {0.49} & {0.21} & 0.12 & {0.20} & {0.31} & 0.26 & 0.01 & 0.37 & {0.36} \\
    SAINT-s           & {0.19} & 0.20 & {0.49} & {0.49} & 0.06 & {0.56} & {0.32} & 0.97 & 0.02 & 0.42 & {0.11} \\
    SAINT-f           & {0.22} & 0.38 & {0.37} & {0.07} & 0.06 & {0.36} & {0.15} & 1.09 & 0.00 & 0.44 & {0.47} \\
    \midrule
    TabDeco           & {0.15} & 0.24 & {0.26} & {0.08} & 0.04 & {0.31} & {0.27} & 0.41 & 0.01 & 0.48 & {0.18} \\
    TabDeco-s         & {0.11} & 0.33 & {0.24} & {0.12} & 0.09 & {0.32} & {0.15} & 0.24 & 0.01 & 0.27 & {0.11} \\
    TabDeco-f         & {0.52} & 0.27 & {0.33} & {0.07} & 0.04 & {0.40} & {0.10} & 0.99 & 0.00 & 0.53 & {0.26} \\
    \bottomrule
    \end{tabular}
    }
    \caption{\small Standard deviations on AUROC scores computed over 10 trials from Table ~\ref{tab: benchmark}. Baseline results are quoted from original papers when possible and reproduced otherwise. Columns denoted by $\dagger$ are multi-class problems, and we report standard errors on accuracy rather than AUROC.}
    \label{tab: variance}
\end{table*}

\end{document}